# Enhancing Layout Hotspot Detection Efficiency with YOLOv8 and PCA-Guided Augmentation

Dongyang Wu, Siyang Wang, Mehdi Kamal, Massoud Pedram
University of Southern California, USA
{wudongya,wangsiya,mehdi.kamal,pedram}@usc.edu

**Abstract**

In this paper, we present a YOLO-based framework for layout hotspot detection, aiming to enhance the efficiency and performance of the design rule checking (DRC) process. Our approach leverages the YOLOv8 vision model to detect multiple hotspots within each layout image, even when dealing with large layout image sizes. Additionally, to enhance pattern-matching effectiveness, we introduce a novel approach to augment the layout image using information extracted through Principal Component Analysis (PCA). The core of our proposed method is an algorithm that utilizes PCA to extract valuable auxiliary information from the layout image. This extracted information is then incorporated into the layout image as an additional color channel. This augmentation significantly improves the accuracy of multi-hotspot detection while reducing the false alarm rate of the object detection algorithm. We evaluate the effectiveness of our framework using four datasets generated from layouts found in the ICCAD-2019 benchmark dataset. The results demonstrate that our framework achieves a precision (recall) of approximately 83% (86%) while maintaining a false alarm rate of less than 7.4%. Also, the studies show that the proposed augmentation approach could improve the detection ability of never-seen-before (NSB) hotspots by about 10%.

## 1 Introduction

In pursuit of increased transistor density and enhanced performance, the semiconductor industry is presently fabricating integrated circuits using smaller feature sizes on advanced process nodes. Notably, contemporary process nodes have the capability to accommodate tens of billions of transistors on a single chip, leading to an unprecedented scale and intricacy of integrated circuits. However, the evolution of process technologies introduces challenges to state-of-the-art circuit design. The reduced feature size at advanced nodes induces a higher probability of defects, necessitating heightened scrutiny during the assessment of designed layout manufacturability. Optical phenomena arising from sub-wavelength features can impede the accurate replication of patterns onto the photoresist, potentially causing undesired disconnections or short circuits in the final manufactured circuit [16]. These incorrectly manufactured patterns are denoted as "hotspots." Consequently, it is imperative to precisely identify and address hotspots in the circuit design, as their mitigation is crucial to averting defects that could compromise the functionality of the circuit.

Traditionally, the circuit printability check was done using the design rule check (DRC) process on the layout. DRC compared the circuit against a large set of design rules to detect violations that would result in hotspots [4]. Pattern matching was also utilized to locate possible hotspots [8, 24, 27]. However, traditional hotspot detection methods are facing significant challenges in current technology nodes. The expanding scale of circuits greatly inflates the time and cost of design and validation, significantly impeding the circuit design flow. New technology nodes also introduce hotspot patterns not seen in former technology nodes, known as the *never-seen-before hotspots* (NSB), undermining the efficacy of locating hotspots in traditional pattern-matching algorithms. Therefore, it becomes imperative to find new and more efficient methods to swiftly perform DRC and accurately identify versatile design hotspots to meet the needs of the current advanced and future technology nodes.

New data-driven methods offer more effective solutions to the hotspot detection problem in the wake of recent progress made in machine learning (ML). ML models loosely describe the collection of algorithms that extract abstract patterns and rules from training data and solve problems using the information obtained from the learning phase [1, 21]. In the hotspot detection application, the common approach is to train the ML model on a set of known layout data with a mix of clips containing or not containing hotspots. The model would first learn to differentiate hotspot and non-hotspot patterns from the labels of the layout data. It would then be used to infer if the to-be-examined layout contains any hotspots. Such a data-driven paradigm is inherently tailored for hotspot detection for two reasons: i) ML models can learn hidden hotspot patterns and detect never-seen-before hotspots while maintaining a swift execution speed, ii) foundries process large quantities of layout data and possess the hotspot information of a given technology, thus making high-quality dataset for training hotspot detection model readily accessible. In addition, since ML models learn abstract hotspot rules directly from training data, they possess the flexibility of continuous optimization and migrating to newer technology nodes by fine-tuning on additional data [26]. These advantages significantly reduce the cost of developing

DRC tools for new technology nodes, making ML the optimal candidate for future circuit hotspot detection applications.

Previously, researchers developed hotspot detection algorithms using multiple ML approaches, including support vector machine (SVM) [19], random forest (RF) [28], and various neural networks (NNs) architecture [2, 5, 6, 11, 14, 15, 25, 26, 29]. Overall, these methods were able to conduct hotspot detection much faster than traditional DRC algorithms [4, 8, 24, 27]. However, the common issues of the previous methods include i) limited or fixed input layout size, ii) lack of multiple hotspot detection and iii) hotspot classification capabilities. On top of these, the model should also achieve a high detection precision and low false alarm rate while maintaining a lightweight size and fast inference speed.

In response to these challenges, we present a YOLO-based hotspot detection framework designed to enhance the efficiency and performance of the DRC process. Our approach leverages the state-of-the-art object detection model, YOLOv8, which has been rigorously tested and proven effective in general object detection tasks, as the foundational element of our framework. The use of object detection DNN models allows us to analyze large layout image sizes and detect multiple hotspots. To further enhance the accuracy and reduce false alarms in hotspot detection, we introduce a novel PCA-based feature extraction method. This method supplements the color channels of the input layout images with auxiliary information, thereby improving the efficacy of pattern matching. To comprehensively evaluate the effectiveness of our proposed framework, we compose four datasets based on ICCAD-2019 [20]. Each dataset includes multiple hotspot and nonhotspot clips within every image to better represent real-world multi-hotspot layout data. We assess the performance of our framework using various metrics, including precision, recall, false positive rate (FPR), F1 score, and runtime.

## 2 Related Work

With the remarkable advances in deep learning witnessed in various domains, the landscape of layout hotspot detection has transitioned towards neural network-based methodologies. Diverse architectural designs have surfaced to tackle the intricate challenges posed by hotspot detection. In this section, we provide an in-depth examination of prior hotspot detection methods, aiming to gain insights into the strengths and limitations inherent to various model architectures.

Baek et al. presented the PGNN model in [2], which combined a convolutional U-Net for extracting routing congestion information with a graph neural network (GNN) responsible for processing the spatial and connectivity aspects of layout pins. While this approach boasted a quicker inference speed, courtesy of its efficient graph representation, it was marred by a relatively higher false alarm rate in comparison to other methods. Also, in [22] a GNN model for detecting single hotspot pattern in the given layout image has been introduced. The proposed GNN model led to a lower false alarm rate. In another vein, an attention-based hotspot detection technique was introduced in [6]. This method displayed commendable accuracy in hotspot detection but grappled with a notable issue of high false alarm rates. Further harnessing the potential of attention mechanisms, Zhu et al. incorporated a transformer encoder in [29] to capture long-range dependencies within layout features. While this method showcased impressive detection performance, it was hampered by the substantial computational costs it incurred.

Convolutional neural networks (CNNs) have emerged as a widely adopted architecture for hotspot detection due to their suitability for processing graphical layout data. In [26], a CNN model tailored for hotspot detection employed graphical layout data, along with extracted pin and route map information, to identify hotspots. A similar CNN-based approach was pursued in [14], albeit with a distinct feature extraction methodology. Reference [5] introduced a modified CNN model called the binarized ResNet, coupled with ensemble learning techniques, to enhance the true positive rate in hotspot detection. However, these CNN models employed fixed input resolutions and could not provide information about the number or specific locations of hotspots. To address the issue of fixed input resolutions and improve hotspot detection flexibility, Xie et al. introduced RouteNet in [25]. RouteNet is a fully convolutional network (FCN) model designed for routing congestion and hotspot detection. It offers the advantage of adaptable input resolutions and can generate hotspot probability maps enriched with spatial information. To further enhance detection performance, [15] introduced the J-Net, a derivative of the FCN architecture, specifically tailored for hotspot detection.

In contrast to previous approaches, we employ an object detection neural network (NN) model to both predict hotspot locations and classify them, all while preserving a lightweight model size and ensuring swift execution speed. Our chosen YOLO-based structure leverages several key features found in prior works, including attention mechanisms and residual networks. A detailed comparison of our proposed hotspot framework with previous methods is presented in Table 1.

## 3 Proposed Method

In this section, we will present the details of our proposed hotspot detector framework. The general flow of the proposed framework is shown in Figure 1, which includes feature extraction and object detection parts. In the following subsections, the details of these two parts are explained.

Table 1. Comparison of major metrics of hotspot detection models

| Work | Model | True Positive Rate | False Positive Rate (Lower is Better) | Input Size | Speed | Multiple Hotspot Detection | Hotspot Classification | Dataset |
|---|---|---|---|---|---|---|---|---|
| [26] | CNN | Low | High | $100 \times 100$ | Low | No | No | ICCAD-2012 |
| [5] | FCN | Low | Low | Flexible | High | Yes | No | ICCAD-2019 |
| [25] | FCN | Low | Medium | Flexible | Low | Yes | No | ISPD-2015 |
| [15] | FCN J-Net | Medium | Medium | Flexible | Medium | Yes | No | In House |
| [14] | CNN | Medium | Low | $224 \times 224 \times 2$ | Medium | No | No | ISPD-2015 |
| [11] | BNN Ensemble | Medium | Medium | NA | Medium | No | No | ICCAD-2019 |
| [2] | GNN U-Net | High | High | Flexible | High | No | No | In House |
| [6] | Attention | High | High | NA | High | No | No | ICCAD-2012 |
| [29] | Transformer | Medium | Low | NA | Low | No | No | ICCAD-2016 |
| Ours | YOLOv8 | High | Medium | Flexible | High | Yes | Yes | ICCAD-2019 |

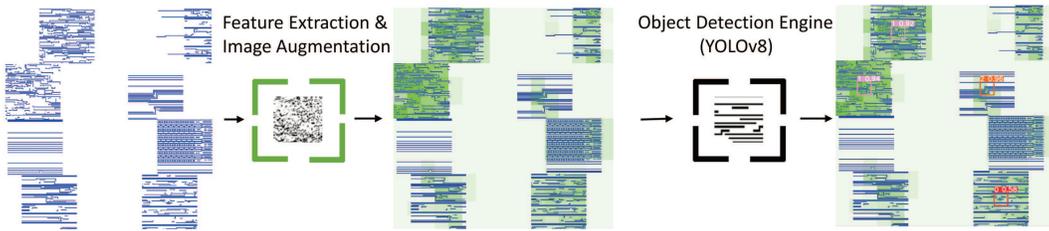

Figure 1. The general flow of the proposed hotspot detector framework

### 3.1 Object Detection

YOLO (You Only Look Once) is a real-time object detection algorithm renowned for its ability to simultaneously predict object bounding boxes and classify objects within those boxes, all in a single forward pass of a neural network. YOLO achieves this by dividing the input image into a grid, with each grid cell responsible for detecting objects within its defined boundaries. For every grid cell, YOLO predicts multiple bounding boxes and their respective class probabilities. Subsequently, it refines these predicted boxes using regression techniques and applies non-maximum suppression to pick the best bounding box. Finally, YOLO generates probabilities for each object class detected within the bounding box. The latest iteration of the YOLO object detection framework is YOLOv8, which introduces a novel ONNX Runtime Supervision architecture, optimizing for both speed and accuracy. It achieves this by incorporating a Transformer backbone instead of traditional Convolutional Neural Networks (CNNs), thereby improving its ability to capture global context.

In the following paragraphs, we will highlight the key attributes of YOLOv8 that position it as a robust choice for the core engine within our hotspot detection framework.

**Large size image compatibility**: When it comes to region-level hotspot detection, YOLOv8 excels in handling large input image sizes. Thanks to its encoder-decoder architecture, the input image size requirements are relatively flexible. Consequently, larger images can be processed without the need for pre-partitioning.

**Data Augmentation**: YOLOv8 employs its own data augmentation technique during training. Throughout the training process, the network introduces random data augmentations, including rotation, flipping, and shearing, to select images. This strategy serves to reduce the risk of overfitting and maintain model stability.

**Comprehensive Feature Analysis**: The backbone of YOLOv8 comprises a PAN-FPN (Path Aggregation Network-Feature Pyramid Network). FPN[17] is capable of generating high-level semantic feature maps by integrating feature maps across various scales. This is useful because convolution tends to enhance features in terms of semantics but reduces their resolution. Additionally, to preserve precise localization information at lower levels, PAN[18], a bottom-up path augmentation is incorporated to enhance low-level information.

**Network Degradation Avoidance**: NNs become more powerful as their depth is increased [10, 13]. However, as depth exceeds a certain point, issues such as vanishing (exploding) gradients[3, 7] can hinder NN convergence and lead to network degradation, resulting in a significant decline in performance. To address this, YOLOv8 incorporates residual networks [9] in the C2f modules, which consist of two convolution layers with residual connections [23]. These residual networks combine their outcomes with the corresponding input, mitigating the risk of network degradation as neural networks become deeper. This approach makes it more practical to use deeper NNs, leading to improved precision in YOLO for object detection tasks. The internal structure of

YOLOv8 is depicted in Figure 2, where the ConvNets within the feature pyramid networks are represented by the term $P$.

In the hotspot detection application, an image of the layout serves as the input to YOLO. YOLO is responsible for predicting the placement of hotspot patterns within the layout and classifying them. In the YOLO model, multiple classes can thus be defined to handle distinct hotspot patterns (and possibly their severity levels.) To train YOLO, a dataset of layout images with annotated hotspot patterns and their corresponding bounding boxes is utilized. The YOLO model is trained with a predefined image size, but, as previously mentioned, it can handle images of arbitrary and larger sizes.

### 3.2 Feature extraction and image augmentation

Hotspot detection presents unique challenges compared to conventional object detection tasks, primarily due to the absence of well-defined boundaries in layout hotspots. This lack of clear boundaries can significantly impact the accuracy of YOLO-based detection. To address this challenge, we introduce an innovative approach to enhance layout images by incorporating extracted features as auxiliary information. Traditional global feature extraction methods like the Discrete Cosine Transform (DCT) may not be well-suited for this task, as they can be influenced by irrelevant background information. Instead, the extracted features should capture the irregularities and local pattern densities within the layout, facilitating YOLO's ability to pinpoint hotspots effectively. To meet this requirement, we propose a Principal Component Analysis (PCA)-based algorithm that extracts information related to local congestion and layout complexity.

In our proposed approach, the image is initially divided into non-overlapping tiles, and our PCA-inspired algorithm is then applied to each tile, generating a new tile with the same dimensions as the feature extracted from the input tile. The pseudocode for the proposed algorithm is provided in Algorithm 1. PCA is a well-established technique used for analyzing and reducing the dimensionality of large datasets, particularly high-resolution images. Its primary objective is to identify orthogonal (uncorrelated) vectors (i.e., vectors with a covariance of 0) as new dimensions where the data is most widely distributed. Geometrically speaking, principal components represent the directions of the data that explain a maximal amount of variance, that is to say, the lines that capture most information of the data. In the application of PCA to 2D layouts, complex layout patterns require more principal components to explain the variance. Therefore, the extracted principal component could reflect the complexity of the layout pattern, which correlates to the presence of hotspots. Our proposed algorithm commences in a manner similar to the PCA algorithm.

When extracting the principal components of a tile (represented as a matrix denoted by $X_0$), we treat each column vector within the tile as a sample. To ensure that each column vector has a mean value of zero, we normalize them by centering the mean value of each column ($M$) to zero. These normalized column vectors are then combined to create a new matrix, denoted as $X$ (line 2). Next, we calculate the covariance matrix $C$ of matrix $X$ (line 3). Subsequently, we perform eigendecomposition on $C$ (line 4) to obtain the corresponding eigenmatrix ($\Lambda$). The eigenmatrix $\Lambda$ should be a diagonal matrix with all elements outside the main diagonal set to 0. The values on the diagonal of $\Lambda$ represent the variances of the data along the eigenbasis of $X$.

---

**Algorithm 1** Proposed Feature Extraction Algorithm

---

**Require:** $X_0 \in \mathbb{R}^{n \times n}, 0 \leq t \leq 1$
**Ensure:** $Y_0 \in \mathbb{R}^{n \times n}$
1. $M \in \mathbb{R}^{1 \times n}, M[i] \leftarrow \overline{X_0[0:n-1][i]}$
2. $X \leftarrow X_0 - \mathbb{1}^{n \times 1} \times M$
3. $C \leftarrow \frac{1}{n} X^T X$
4. Find $E, \Lambda \in \mathbb{R}^{n \times n}$ s.t.:
   $(C = E \Lambda E^{-1}) \wedge (\Lambda[i][j] \times (i-j) = 0, \forall i, j \in [0, n-1]$

5. $B_0[i] \leftarrow \Lambda[i][i]$, sort $B_0$ s.t.:
   $(B_0[i] - B_0[j]) \times (i-j) \leq 0, \forall i, j \in [0, n-1]$
6. $B \leftarrow B_0[0:k-1]$ s.t.:
   $sum(B) \geq t \times sum(B_0) > sum(B[0:k-2])$
7. $Y_0[i][j] \leftarrow k, \forall i, j \in [0, n-1]$

---

Conventional PCA algorithms typically achieve image compression by discarding eigenbases that are responsible for insignificant variances. In contrast, our proposed algorithm defines the extracted feature as the minimum number of eigenbases required to capture a predetermined amount of sample variance, denoted as the threshold value $t$. This threshold value is critical as it reflects the complexity of the local layout pattern. Line 5 of the algorithm extracts the variance values and stores them in the vector $B_0$, subsequently sorting them in descending order. We then extract $k$ largest variance values whose cumulative sum exceeds the threshold $t$ percentage of the sum of all variances, which is stated in Line 6. Finally, the number of extracted variances, represented by the variable $k$, is determined in Line 7 of the algorithm. The value $k$ serves as a compact representation of the entire tile's information. In the augmented RGB image of the layout clip, the pixels in the second channel (green) are set to the $k$ value corresponding to their respective tiles. Pixels in the last channel retain the original layout information, while the first channel remains unused, with all pixels set to 0.

Figure 3 illustrates the corresponding extracted features for two ICCAD-2019 layouts [20]. As depicted in the figure, distinct patterns within the tiles result in different extracted feature values, whereas similar patterns yield identical feature values (refer to Figure 3(b) and (c)). It is important to point out that for the layout in Figure 3(d), the tile positions and the potential slight variations in their patterns result in highly similar, although not identical, extracted feature values for all tiles. Note that the background of the layouts in the figures has been altered from black to white to enhance visual clarity.

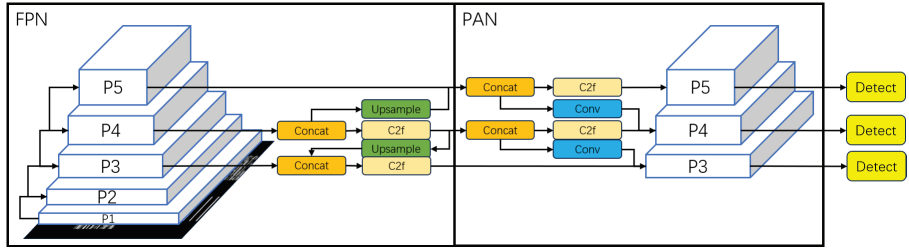

**Figure 2.** The internal structure of YOLOv8.

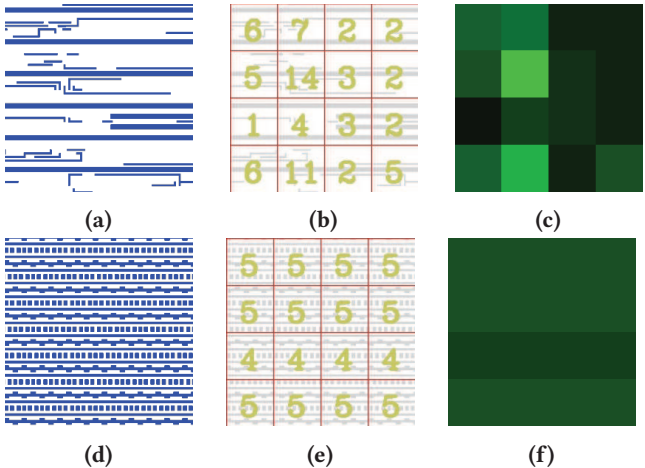

**Figure 3.** (a),(d) The two ICCAD-2019 layouts (clips). (b),(e) The tiles and the number of chosen variance values (variable $k$ in Algorithm 1) for the input image layouts when parameter $t$ was defined as 99.9%, and (c),(f) the corresponding feature image of the input layouts.

A key parameter in the proposed feature extraction algorithm is the tile size, which determines the characteristic size of the extracted features. Through our observations, we have noticed that certain layout clips contain patterns that appear simple at a smaller scale but form intricate structures when examined at a larger scale. Conversely, some clips exhibit the opposite behavior, where patterns appear complex at a smaller scale but exhibit regular arrangements when viewed at a larger scale. To accommodate these variations, we recommend extracting features using two different tile sizes. The larger tile size should be four times the size of the smaller one. The final extracted feature for the layout is computed as the weighted average of these two distinct features.

## 4 Results and Discussion
### 4.1 Experimental Setup

To demonstrate the efficacy of our method, we opted for the ICCAD-2019 dataset, which includes complex hotspot patterns as well as never-seen-before (NSB) hotspots, as the foundation for our dataset. The training dataset of ICCAD-2019 consists of a variety of layout clips, some containing hotspots at the center and others without hotspots, along with their corresponding hotspot labels. To evaluate the multiple-hotspot detection capability of our framework, we created new layouts by combining multiple clips from ICCAD-2019 onto larger canvases. Since the clips composing the new layout images are sampled randomly from ICCAD-2019, the difficulty of hotspot detection is increased compared to the original one.

We created four datasets, each comprising images with dimensions of 1024 × 1024 pixels. Initially, we resized each clip from ICCAD-2019 to a 256 × 256-pixel image using Klayout Python Macro programming [12]. Datasets 1 and 4 were then formed by randomly arranging clips selected from ICCAD-2019 into a mosaic layout with regular rows and columns. Dataset 1 contains 16 clips (arranged in 4 rows and 4 columns) within each image, while Dataset 4 contains 64 clips (arranged in 8 rows and 8 columns). To ensure that the images remained 1024 × 1024 pixels in size, each chosen clip was resized to 128 × 128 pixels. Dataset 2 consisted of images featuring four rows and two columns of clips, unlike the regular arrangement seen in Datasets 1 and 4. In Dataset 2, the horizontal placement of clips within each image was randomized to prevent alignment with clips from the same column. Dataset 3 adopted a placement approach similar to Dataset 2 but with the addition of sparsely positioned wires that connected clips within the same row. These wires were introduced using a routing scheme designed to reflect more realistic layouts without introducing new hotspots.

A sample from each dataset is depicted in Figure 4. Each dataset consisted of 10,000 training samples and 2,000 test samples, with the test data generated from the ICCAD-2019 Testing dataset-I benchmark. Table 2 provides details on the number of hotspot and non-hotspot clips within each dataset. The hotspots were identified as a central area measuring 64 × 64 pixels within hotspot clips. We further categorized these hotspots into 4 classes based on their corresponding benchmark indices in the ICCAD-2019 dataset. To assess the sensitivity of the trained model to layout size scaling, we employed the models trained on Dataset 1 for validation purposes in the case of Dataset 4.

**Table 2.** The features of the generated datasets (HS: hotspot, NHS: nonhotspot).

| Dataset | Training Data | | Testing Data | |
|---|---|---|---|---|
| | HS # | NHS # | HS # | NHS # |
| 1 | 45006 | 114994 | 8957 | 23043 |
| 2 | 5060 | 54940 | 5000 | 11000 |
| 3 | 24970 | 55030 | 5009 | 10991 |
| 4 | 84609 | 555391 | 16926 | 111074 |

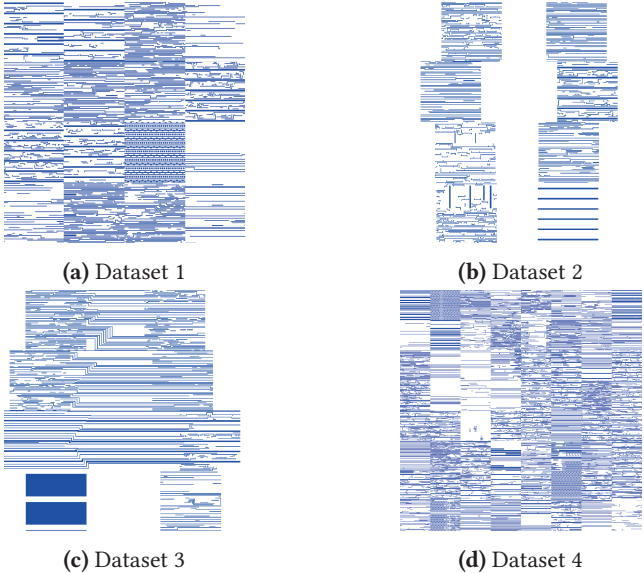

(a) Dataset 1   (b) Dataset 2

(c) Dataset 3   (d) Dataset 4

**Figure 4.** Sample layout images from generated datasets.

For our YOLOv8 model, we utilized the Ultralytics YOLOv8 package. Among the various available sizes of YOLOv8, we opted for the smallest one (YOLOv8n) with 3.2 million parameters due to its faster inference and training speed. Our hotspot detection framework was fully implemented in Python and accelerated using an NVIDIA GeForce RTX 2080 GPU for both training and testing. It's worth noting that the feature extraction algorithm ran on the CPU (Intel i9 with 3.5GHz operating frequency), while the YOLO algorithm utilized the GPU. During model training, we set the initial learning rate to 0.01, and the maximum training epoch was defined as 100.

We used metrics defined as the following to measure the performance of our proposed framework:

- *Precision*: The ratio between the number of correctly detected hotspots and the total number of predictions.
- *Recall*: The ratio between the number of correctly detected hotspots and the total number of ground-truth hotspots.
- *F1 Score*: The harmonic mean of precision and recall.

$$F1 = \frac{2 \times Precision \times Recall}{Precision + Recall} \quad (1)$$

Here, the accurate detection is characterized by an intersection between the ground truth box and the prediction box that encompasses at least 50% of their combined area.

### 4.2 Discussion

We present the evaluation metrics for the proposed framework, averaged across the four hotspot classes, in Table 3. The results demonstrate the robust hotspot detection capabilities of our YOLOv8-based framework. Across Dataset 1-3, our method achieved impressive precision and recall, averaging around 83% and 85%, respectively, while maintaining a low false-positive rate (FPR) at approximately 5%. In a comparative study, we noticed a decrease in detection accuracy on Dataset 4. We attribute this performance drop to the difference in image sizes between the model trained on 256 × 256 clips and the downsized 128 × 128 test inputs used in Dataset 4. The total execution time for YOLOv8n on all samples in each dataset was up to 20 seconds. Thus, the detection runtime of the proposed approach (without feature extraction overhead) was about 5ms for each image (0.3ms for each clip), which is lower than that of most prior works (e.g., [26] and [11]). However, the runtime of the proposed method of [5] (on Nvidia P100 GPU and without feature extraction overhead) was about 0.6ms for each clip of dataset-I of ICCAD-2019 with the size of 105 × 105 pixels. The feature extraction process runtime of our proposed method was about 10ms on the CPU for each clip.

Figure 5 illustrates the precision and recall of our framework for different classes in Dataset 1. The most challenging class is class 1, where hotspots are generated by closely parallel lines, making it difficult for object detection algorithms to distinguish them from non-hotspot patterns. ICCAD-2019 comprises four categories of hotspot classes, with the initial class featuring a substantial collection of never-seen-before hotspots. Our research indicates that employing PCA predominantly enhances the efficiency of YOLO in detecting this specific hotspot type, resulting in an approximately 10% improvement in recall.

**Table 3.** The performance metrics of the proposed hotspot detection framework.

| Dataset | Precision | Recall | F1 | FPR |
|---|---|---|---|---|
| 1 | 83.2% | 86.9% | 85.0% | 7.4% |
| 2 | 85.3% | 85.8% | 85.5% | 4.5% |
| 3 | 90.7% | 82.5% | 86.4% | 3.7% |
| 4 | 72.3% | 84.0% | 77.7% | 4.5% |

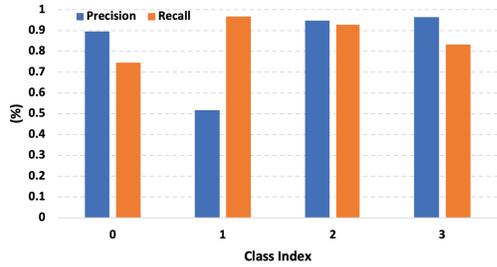

**Figure 5.** The precision and recall of the proposed framework for different classes in Dataset 1.

The performance metrics of prior work evaluated on testing dataset-I of ICCAD-2019 are summarized in Table 4. Note that the results from [26] were obtained from the reported values in [5]. The comparison highlights the superior performance of our proposed framework in terms of precision, recall, and F1-score, which are consistently higher than those of the prior works. On average, our framework achieves approximately 14% higher precision, 11% higher recall, and a remarkable 36% higher F1-score compared to the prior work. However, it's important to note that our false positive rate is slightly higher, with a maximum difference of 4% compared to the prior work. A noteworthy consideration is that each test sample in the prior work consisted of a single clip from the ICCAD-2019 benchmark, transforming the hotspot detection issue into an image classification challenge. However, our approach employed samples containing at least six clips (datasets 2 and 3), where our proposed approach was tasked with identifying the position and categorizing the hotspot patterns. This increased dataset complexity likely contributed to the differences in performance, highlighting the effectiveness of our approach in handling more intricate layouts.

**Table 4.** The performance metric of the prior work which has been evaluated on testing dataset-I ICCAD-2019.

| Work | Precision | Recall | F1 | FPR |
|---|---|---|---|---|
| [26] | 66.7% | 76.0% | 71.0% | 2.6% |
| [11] | 68.9% | 80.9% | 74.4% | 2.5% |
| [5] | 61.0% | 69.1% | 64.8% | 3.0% |

## 5 Conclusion

In this paper, we presented a framework for detecting and classifying the hotspots of the layouts by employing the YOLOv8 model. Also, we augmented the layout images, with auxiliary information to improve the accuracy of the detection process. To extract this information, we suggested a feature extraction algorithm based on the PCA. The studies on four datasets which have been generated by employing the layouts of the ICCAD-2019 benchmark dataset showed the high efficacy of the proposed framework.